%% file: main.tex
\documentclass[runningheads]{llncs}

\usepackage{eccv}

\usepackage{eccvabbrv}

\usepackage{graphicx}
\usepackage{multirow}
\usepackage{booktabs}

\usepackage[accsupp]{axessibility}  

\newcommand{\name}{$\mathtt{HybridBooth}$}

\usepackage[pagebackref,breaklinks,colorlinks,citecolor=eccvblue]{hyperref}
\usepackage{hyperref}

\usepackage{orcidlink}

\usepackage{marvosym}

\begin{document}

\title{HybridBooth: Hybrid Prompt Inversion for Efficient Subject-Driven Generation} 
\titlerunning{HybridBooth}


\author{
Shanyan Guan\inst{1}\thanks{~Equal contributions. Ying Tai and Mingyu You are corresponding authors. }\orcidlink{0000-1111-2222-3333} \and
Yanhao Ge\inst{1,3}$^{\star}$\orcidlink{0000-0002-5650-5118} \and 
Ying Tai\inst{2}\textsuperscript{\Letter}\orcidlink{0000-0002-4665-6852} \and
Jian Yang\inst{2} \\
Wei Li\inst{1} \and
Mingyu You\inst{3}\thanks{Mingyu You is also affiliated with State Key Laboratory of Intelligent Autonomous Systems, Frontiers Science Center for Intelligent Autonomous Systems, Shanghai Key Laboratory of Intelligent Autonomous Systems}
\textsuperscript{\Letter}\orcidlink{0000-0003-2758-167X}
}

\authorrunning{S. Guan, Y.~Ge, Y.~Tai, J.~Yang, W.~Li, and M.~You}

\institute{
$^1$~vivo Mobile Communication Co., Ltd \\
$^2$~School of Intelligence Science and Technology, Nanjing University \\
$^3$~College of Electronic and Information Engineering, Shanghai Research Institute for Intelligent Autonomous Systems, Tongji University\\
\email{\{guanshanyan, halege\}@vivo.com} ~~ \email{yingtai@nju.edu.cn} ~~ \email{myyou@tongji.edu.cn} \\
\url{https://sites.google.com/view/hybridbooth}
}

\maketitle

\input{camera_ready/abs}

\input{camera_ready/intro}

\input{camera_ready/related}

\input{camera_ready/method}
\input{camera_ready/exp}

\input{camera_ready/concl}


%
%
\bibliographystyle{splncs04}
\bibliography{main}
\end{document}

%% file: camera_ready/abs.tex
\begin{abstract}

Recent advancements in text-to-image diffusion models have shown remarkable creative capabilities with textual prompts, but generating personalized instances based on specific subjects, known as subject-driven generation, remains challenging. To tackle this issue, we present a new hybrid framework called \name, which merges the benefits of optimization-based and direct-regression methods. \name~operates in two stages: the Word Embedding Probe, which generates a robust initial word embedding using a fine-tuned encoder, and the Word Embedding Refinement, which further adapts the encoder to specific subject images by optimizing key parameters. This approach allows for effective and fast inversion of visual concepts into textual embedding, even from a single image, while maintaining the model's generalization capabilities. 

\keywords{Customized Generation \and Image Inversion \and Image synthesis}
\end{abstract}

%% file: camera_ready/intro.tex
\section{Introduction}\label{sec:intro}

\begin{figure}
\begin{center}
    \centering
    \includegraphics[width=0.89\linewidth]{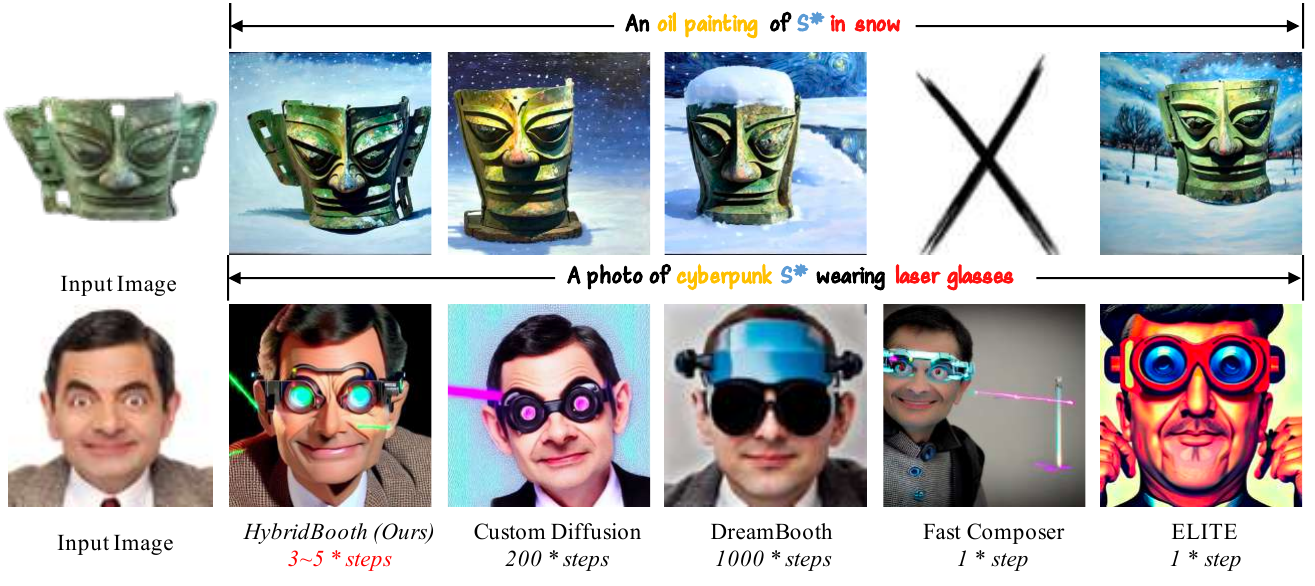}
    \caption{We propose~\name~for efficient subject-driven generation. Unlike FastComposer~\cite{xiao2023fastcomposer}, which targets humans, it cannot affect the Bronze Mask (Top). Compared to optimization-based~\cite{kumari2022customdiffusion, dreambooth} and direct-regression-based methods~\cite{xiao2023fastcomposer, wei2023elite}, our framework generates diverse, precise images while preserving the subject's identity, requiring only $3\sim5$ optimization iterations.}
    \label{fig:fig1}
\end{center}%
\end{figure}

Recent advancements in text-to-image diffusion models~\cite{ho2020denoising,SD} have demonstrated unprecedented creating capabilities guided by textual prompts, such as creative endeavors in visual arts~\cite{pan2022synthesizing}, synthetic dataset build-up~\cite{voetman2023big}, and image editing for marketing and advertising~\cite{mokady2022null}.
However, creating personalized instantiations based on specific subjects remains a challenging problem, also known as subject-driven generation~\cite{dreambooth, textualinversion, wang2024instantid,gal2023e4t,peng2024portraitbooth}. 
 
To achieve subject-driven generation, recent works~\cite{dreambooth, textualinversion, gal2023e4t} explore the inversion of subject images into specific textual embeddings, either through iterative optimization (optimization-based) or direct regression methods. These two paradigms are summarized in~\cref{fig2:comparison}.
Optimization-based methods, such as Textual Inversion and DreamBooth~\cite{textualinversion, ruiz2023hyperdreambooth}, iteratively optimize the generation model or textual embedding of the subject to better align with given subject images. Constrained by tens of subject images, optimization-based methods can accurately capture the identifying characteristics of the subject. However, the optimization procedure is \textit{computationally expensive} and \textit{slow to converge}.
In comparison, direct-regression-based methods~\cite{wei2023elite, gal2023e4t, arar2023domainagnostic, tov2021designing, shen2020interfacegan, shen2020interpreting, choi2018stargan, alaluf2021ageregression, alaluf2023NeTI, Tewel2023KeyLockedRO} achieve textual concept inversion by tuning a pre-trained model with labeled datasets or concept-specific images. While these methods enable zero-shot subject-driven generation, they suffer from \textit{detail loss and adaptation issues} when combining the target subject with new styles or prompts.

In this paper, we introduce a novel framework called \name, designed for efficient subject-driven generation using text-to-image diffusion models. The core idea is to synergistically combine the strengths of various methods, creating a hybrid approach that balances quality and speed.
As shown in \cref{fig2:comparison}, \name~operates in two stages. The first stage, termed Word Embedding Probe, employs an encoder to generate a robust initial word embedding. This encoder is fine-tuned on a large-scale dataset such as FFHQ~\cite{ffhq} with 70,000 images.
In the second stage, the domain-agnostic encoder undergoes further fine-tuning to adapt to a specific subject's image in a word embedding refinement process. This stage focuses on optimizing the most influential parameters of the regressor.
These two stages work collaboratively, enhancing each other to achieve precise and efficient subject-driven generation.

Overall, our contributions are summarized as follows:
\begin{itemize}
        \item Combining strengths of both optimization-based and direct-regression-based paradigms, we propose \name~for precise and efficient subject-driven generation in less than $5$ iteration steps. 
        \item We craft a Regressor-based Word Embedding Probe module to ensure a robust feature initialization while maintaining a focus on general-oriented generation model.
        \item We introduce a residual refinement strategy to accelerate the fitting process while preventing the collapse of the prior learned during the probe stage.
\end{itemize}

%% file: camera_ready/related.tex
\section{Related Work}

\subsection{Text-to-image Generation}

Text-to-image generation focuses on creating visually accurate and contextually relevant images from textual descriptions. Early approaches, such as Generative Adversarial Networks (GANs)~\cite{gan, cyclegan2017, Pix2Pix2017, karras2019stylegan}, achieved high-quality synthetic images. However, these studies primarily emphasize image generation within specific domains, leading to limitations in extended applications.
The progress of large language models~\cite{Devlin2019BERTPO, clip, brown2020language, touvron2023llama, touvron2023llama2} has enabled methods based on variational auto-encoders~\cite{vqvae, kingma2022vae} and diffusion models~\cite{SD, saharia2022Imagen, ramesh2021dalle} to effectively map word sequences into latent space, guiding image generation from textual prompts.
Our method leverages pre-trained text-to-image diffusion models to generate personalized visual content efficiently and effectively.

\begin{figure}[t]
    \centering
    \includegraphics[width=0.9\linewidth]{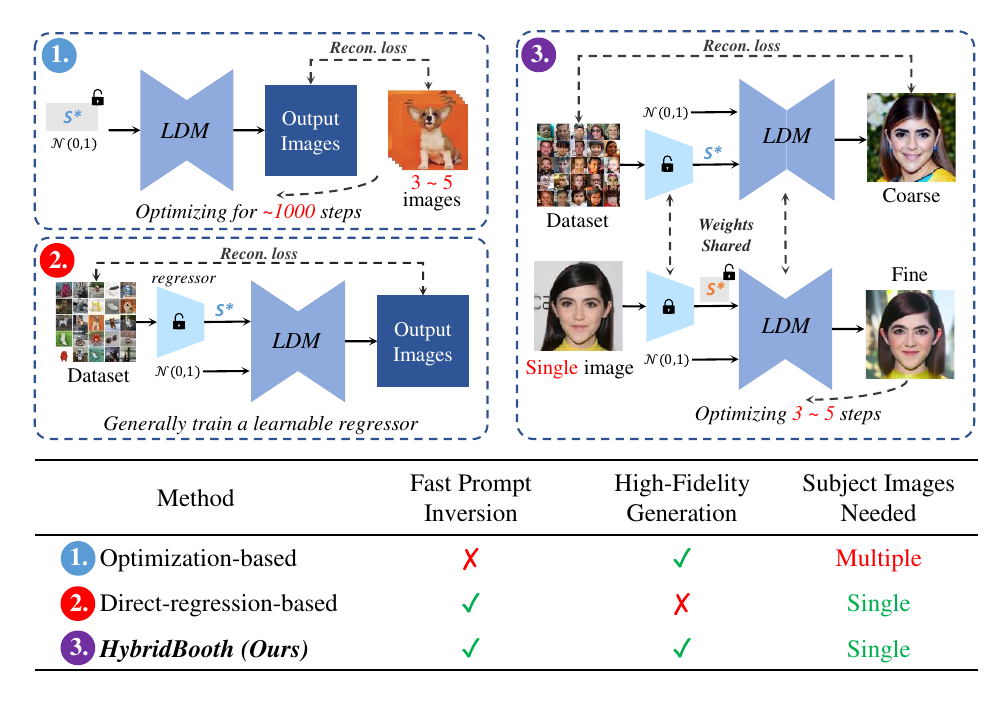} 
    \caption{Comparing with previous frameworks, our method leverages a coarse-to-fine strategy, effectively integrating the advantages of both optimization- and regression-based methods. Consequently, we enable a \textit{fast prompt inversion }with \textit{high-fidelity generation} from just a \textit{single} image.}
    \label{fig2:comparison}
\end{figure}

\subsection{Subject-driven Generation}~\label{sec:subject_related}

Subject-driven generation is a type of controllable generation task that aims to generate instances of specific subjects in new contexts (\eg, styles, scenarios). Current approaches can be broadly categorized into two types: 1) optimization-based and 2) direct-regression-based methods.

\paragraph{Optimization-based Methods} iteratively refine images to align with given textual descriptions. Inversion is a common technique to preserve a subject while modifying its context. For GANs, Pivot Tuning~\cite{roich2021pivotal} fine-tunes the model using an inverted latent code for robust, ID-preserving generation. Recent work~\cite{pan2023dragGAN} keeps GAN-generated subjects consistent and controllable. However, these methods often struggle with unique subjects and fail to preserve all details. In diffusion models, inversion aims to determine the initial noise that generates the target image through denoising. Methods like DreamBooth and Textual Inversion~\cite{dreambooth, textualinversion, ruiz2023hyperdreambooth, kumari2022customdiffusion} personalize pre-trained models to generate specific subjects in various contexts but are computationally intensive and slow to converge. Our method uses a hybrid framework for robust initial feature encoding, followed by rapid optimization, achieving superior efficiency.

\paragraph{Direct-regression-based Methods} generate subject-specific images directly from textual prompts without iterative optimization, relying on extensive model modifications on large datasets. These methods generate images from scratch based on learned text-visual correlations. Previous GAN-based methods~\cite{tov2021designing, shen2020interfacegan, shen2020interpreting, choi2018stargan, alaluf2021ageregression} retrain generators or add conditioned constraints for ID-preserving generation but are limited by their training sets. Recent diffusion-based methods~\cite{gal2023e4t, wei2023elite, arar2023domainagnostic, alaluf2023NeTI, Tewel2023KeyLockedRO} tune only the encoder of pre-trained models, offering efficiency but risking overfitting and reduced performance on diverse tasks. To address these limitations, \name~initially refines the encoder and employs a subject-driven residual refinement module. This process requires only 3 to 5 steps, resulting in more detailed and robust images compared to other direct-regression-based methods.

%% file: camera_ready/method.tex
\section{Preliminaries}

\subsection{Text-to-Image Latent Diffusion Models}
We implemented our method based on text-to-image Latent Diffusion Models (LDMs)~\cite{SD} which can understand prompts and generate vivid images. 
LDMs mainly consist of three components, including a variational auto-encoder (VAE)~\cite{vqvae}, an UNet, and a CLIP text encoder~\cite{clip}. 
The VAE encoder $\mathcal{E}$ compresses $\boldsymbol{x}$ into a low-dimensional latent code $\boldsymbol{z} = \mathcal{E}(\boldsymbol{x})$, and the decoder $\mathcal{D}$ maps the latent code $\boldsymbol{z}$ back to image space $\boldsymbol{\hat{x}} = \mathcal{D}(\boldsymbol{z})$. 
For the text-guided diffusion process, the latent code $\boldsymbol{z}$ is perturbed by Gaussian noise $\boldsymbol{\epsilon}$. Then the UNet $\mathcal{M}_{\boldsymbol{\theta}}$ learns to predict the applied noise $\boldsymbol{z}_t$ with a specified time step $t$. The denoising process is conditioned on the text feature $\boldsymbol{c}$ extracted from text prompts $\boldsymbol{y}$ by using text encoder $\tau_{\boldsymbol{\theta}}$. 
The training loss is formulated as:
\begin{align}
    \mathcal{L}_{\boldsymbol{\epsilon}} = \mathbb{E}_{\boldsymbol{z}\sim\mathcal{E}(\boldsymbol{x}), \boldsymbol{\epsilon}\sim\mathcal{N}(0,1), \boldsymbol{c}, t} \left[\big\| \boldsymbol{\epsilon} - \mathcal{M}_{\boldsymbol{\theta}}(\boldsymbol{z}_t, \boldsymbol{c}, t) \big\|_{2}^{2}\right].
    \label{eq:ldmloss}
\end{align}
During inference, given a prompt $\boldsymbol{y}$, the model iteratively denoises $\boldsymbol{z}_t$ conditioned on the text feature $\boldsymbol{c}=\tau_{\boldsymbol{\theta}}(\boldsymbol{y})$. Finally, $\boldsymbol{z}_0$ is decoded to an image $\hat{\boldsymbol{x}} = \mathcal{D}(\boldsymbol{z}_0)$.

\begin{figure}[t]
    \centering
    \includegraphics[width=\linewidth]{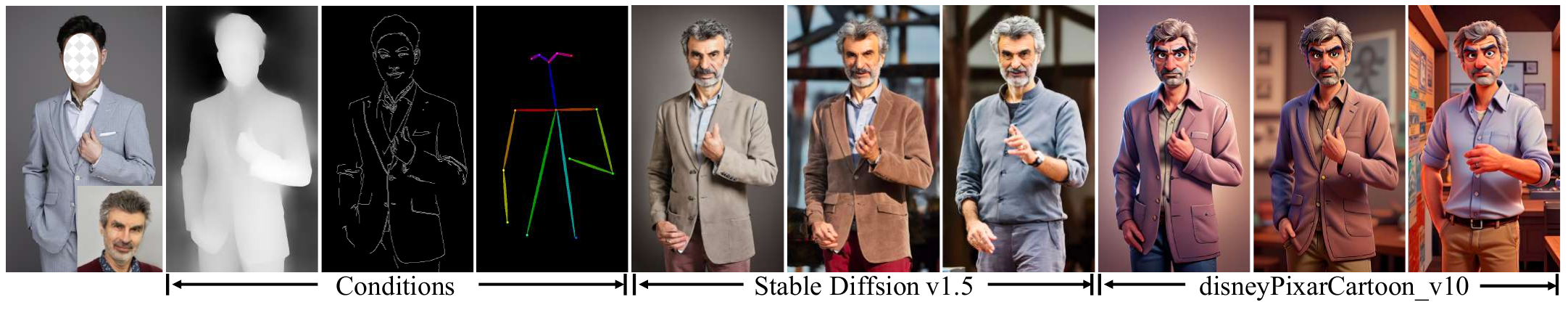}
    \caption{Showcasing our prompt inversion framework incorporating various conditions (depth, canny, keypoints) and community models\protect\footnotemark.}
    \label{fig:control_control}
\end{figure}
\footnotetext{disneyPixarCartoon\_v10: civitai.com/models/65203/disney-pixar-cartoon-type-a}

\subsection{Prompt Inversion for Subject-Dirven Generation}\label{sec:subject_driven_generation}
Despite the remarkable general generation capability of LDMs, challenges remain in generating specific subjects across various contexts, known as Subject-Driven Generation~\cite{dreambooth, textualinversion}.
One feasible approach is to learn textual concepts from subject-specific images, referred to as Prompt Inversion~\cite{gal2023e4t,blip-diffusion,arar2023domainagnostic}.
Given several images $\boldsymbol{X}$ of a subject, the prompt inversion aims to inverse the visual content of the subject to the embedding of words $\boldsymbol{e}$:
\begin{align}
    \boldsymbol{e} \leftarrow \mathcal{I}(\boldsymbol{X}),
\end{align}
where $\mathcal{I}$ represents the inversion process. The primary goal of the inversion process is to ensure that $\boldsymbol{e} \in \mathbb{R}^{n\times d}$ captures the key identifying characteristics of the subject.  The word embedding $\boldsymbol{e}$ is then encoded into textual features $\boldsymbol{c}$ using a pre-trained text encoder (\ie, CLIP).
Once $\boldsymbol{e}$ is learned, it can be readily combined with arbitrary text for the creative generation of the subject. Specifically, let `S*' be the text word corresponding to the learned embedding $\boldsymbol{e}$, we can combine `S*' with new prompts to realize arbitrary subject-driven image synthesis, \eg `S* in snow forest'.

\paragraph{Discussion.} Another paradigm involves mapping subject images to image tokens and using an additional cross-attention layer to inject these tokens into the diffusion process. This paradigm significantly alters the generation process of the original Latent Diffusion Models (LDMs), necessitating retraining on millions of images. In contrast, prompt inversion is a more resource-efficient method for subject-driven generation. As shown in \cref{fig:control_control}, it can seamlessly integrate with various community foundational models and controllers pre-trained on the original stable diffusion, making it a versatile and practical alternative.

\begin{figure}[t]
    \centering  \includegraphics[width=0.9\linewidth]{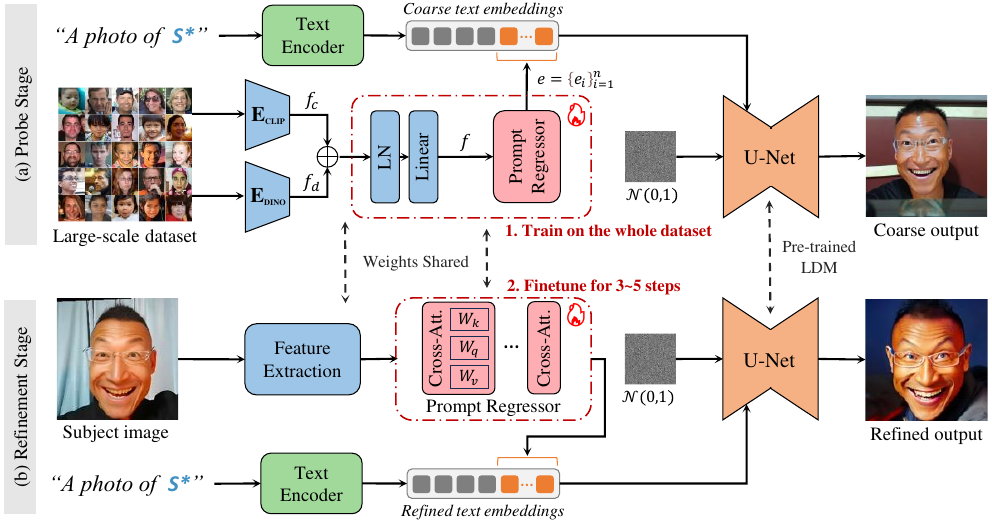}
    \caption{The framework of~\name~consists of two stages: (a) The first \textit{Probe Stage} leverages a prompt regressor estimate the initial word embedding, acting as a probe for a rational word embedding; (b) The second \textit{Refinement Stage} adapts the regressor to the subject by taking only 3-5 iterations to achieve efficient subject generation.
    }
    \label{fig:framework}
\end{figure}

\section{Hybrid Prompt Inversion Framework}
\label{sec:framework}

Given just \textit{one} image $\boldsymbol{x}$ of a specific subject, we aim to efficiently embed the subject into the word embedding space. 
We combine the strengths of optimization- and direct-regression-based paradigms to achieve precise and efficient subject-driven generation. 
Thus, we propose a hybrid prompt inversion framework, termed~\name. 
As shown in \cref{fig:framework}, \name~consists of two stages: 
($1$) \textbf{Word Embedding Probe Stage}, which learns a domain-agnostic encoder providing an initial word embedding estimate.
($2$) \textbf{Word Embedding Refinement Stage}, which adapts the encoder to the given image of the subject. 
The key idea is: \textit{A powerful encoder reduces iterative refinement costs}, while \textit{effective iterative refinement reduces the encoder's accuracy demands}. 
Our challenge is to develop a robust encoder for initial embeddings and an effective refinement scheme to reduce the encoder's precision requirements.
Next, we introduce the Word Embedding Probe stage in \cref{sec:probe}, describe the residual refinement stage in \cref{sec:refinement}, and present the framework details in \cref{sec:implementation}.

\subsection{Regressor-based Word Embedding Probe}
\label{sec:probe}

First,~\name~trains a prompt regressor that takes an image of a subject as input and estimates the initial word embedding. This involves extracting multi-grained image features and conducting multiple-word regression.

\paragraph{Multi-grained Image Feature Merging.} 
Existing regressor-based methods exclusively utilize CLIP features $\boldsymbol{f}_{c}$, which often contain global semantic information (\eg, subject's category)~\cite{jiang2023clip, clip, dreambooth}. 
Hence, we incorporate DINOv2~\cite{oquab2023dinov2} features $\boldsymbol{f}_{d}$ as a supplement, offering detailed pixel-level information.
Taking inspiration from previous practices~\cite{jiang2023clip, xiao2023fastcomposer}, we employ a Multilayer Perceptron (MLP) to project DINOv2 features $\boldsymbol{f}_{d}$, then concatenate the output with CLIP feature $\boldsymbol{f}_{c}$. Finally we align the feature dimension using a Linear layer~\cite{rosenblatt1958perceptron}. The merging process is formulated as:
\begin{align}
    \boldsymbol{f} = \text{Linear}\left([\boldsymbol{f}_{c}, \text{MLP}(\boldsymbol{f}_{d})]\right),
\end{align}
where $[\cdot, \cdot]$ represents the concatenation operation, and the structure of MLP is \texttt{LyaerNorm-Linear-GELU~\cite{zhang2018gelu}-Linear}. 
As verified in our experiments, this simple module yields significant improvements.

\paragraph{Multiple-word Regression.}
The merged feature $\boldsymbol{f}$ is then mapped to word embeddings $\boldsymbol{e}$.
Typically, a single word represents a specific concept, \eg `man' conveys the subject's category but may lack details like hairstyle. 
Thus, we suggest projecting the feature $\boldsymbol{f}$ to multiple-word embeddings to provide a more thorough description of the subject:
\begin{align}
    \boldsymbol{e} = \mathcal{R}(\boldsymbol{f}), 
\end{align}
where $\boldsymbol{e} = \{\boldsymbol{e}_{i}\}_{i=1}^{n}$, $n$ is the number of word embedding $\boldsymbol{e}_{i} \in \mathbb{R}^{d}$, and $d$ is the dimension of word embedding. 

\begin{figure}[t]
    \centering
    \includegraphics[width=0.95\linewidth]{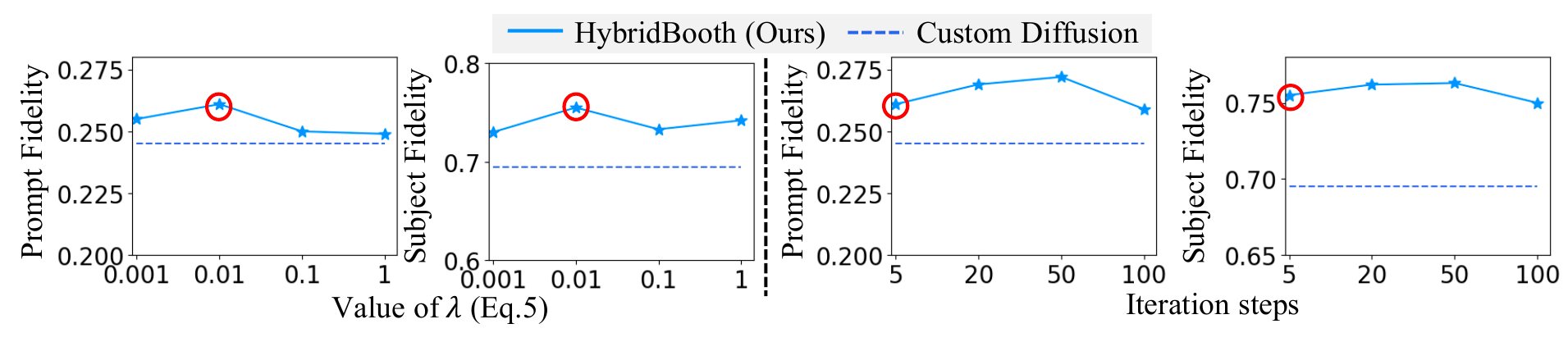}
    \caption{
    Impact of hyper-parameters in the residual refinement stage (on the DreamBooth dataset~\cite{dreambooth}). Default parameters are marked with red circles. Custom Diffusion~\cite{kumari2022customdiffusion} serves as the reference. Prompt Fidelity and Subject Fidelity are CLIP-I and DINO-T, respectively (see \cref{sec:exp_detail}).
    }
    \label{fig:hyperparam}
\end{figure}

\subsection{Residual Refinement}
\label{sec:refinement}
Although the regressor-based probe is fast, precise alignment between word embeddings and key subject features remains challenging without image feature feedback. 
Thus, our subsequent stage focuses on refinement, aiming to adapt the learned regressor to target subject for better alignment. 
Striking a balance is challenging due to the \textit{single} image input, as it requires avoiding overfitting to a specific image while capturing essential subject characteristics effectively.

To tackle this issue, we enhance the learned regressor with a residual refinement strategy that adjusts only critical parameters ($\boldsymbol{W}_{\boldsymbol{\phi}}$) for faster adaptation while preserving the learned prior. 
The residual refinement is formulated as:
\begin{align}
    \boldsymbol{W}_{\boldsymbol{\phi}}^{'} = \boldsymbol{W}_{\boldsymbol{\phi}} + \lambda \Delta \boldsymbol{W}_{\boldsymbol{\phi}},
    \label{eq:partial_refine}
\end{align}
where $\Delta \boldsymbol{W}_{\boldsymbol{\phi}}$ is the learned residual parameters and a hyperparameter $\lambda$ is to control the scale of $\Delta \boldsymbol{W}_{\boldsymbol{\phi}}$. 
With residual refinement, it only takes $3$-$5$ iterations to achieve high-fidelity subject generation while preserving editability.

\paragraph{Discussing the overfitting problem.} 
Overfitting is a common challenge in finetune-based methods, especially when tuning the prompt regressor on a single image. To mitigate this, we refine only \textit{partial} parameters (see \cref{eq:partial_refine}). 
Intuitively understanding, $\boldsymbol{W}_{\phi}$ acts as an anchor, preserving the learned prior and stabilizing updates. Zeroing out $\boldsymbol{W}_{\phi}$ collapses its prior (see the last column in \cref{fig:framework_ablation}), and we need to adjust hyper-parameters more carefully.
As shown in \cref{fig:hyperparam}, by using residual refinement, $\lambda$ and iteration steps can vary widely yet still yield good results. 
\cref{fig:comparison} and \cref{fig:general_obj} demonstrate that \name can generate novel subject images with diverse expressions, postures, styles, and scenarios.

\subsection{Framework Details}
\label{sec:implementation}

\paragraph{Prompt Regressor Design.}
Previous studies have used either an MLP or a UNet as the prompt regressor. \name~is flexible and can integrate any network mapping image features $\boldsymbol{f}$ to word embeddings $\boldsymbol{e}$. We employ the state-of-the-art regressor PromptNet~\cite{zhou2023enhancing}, which uses a block structure similar to LDM and can be initialized with LDM pretrained on large-scale image datasets.
In the refinement stage, we focus on refining the cross-attention layer parameters (key, query, and value matrices $\boldsymbol{W}_{k}$, $\boldsymbol{W}_{q}$, $\boldsymbol{K}_{v}$) based on layer importance experiments. Fine-tuning the entire regressor with $100$ images, we calculated importance scores for each layer based on the mean cosine distance of parameters before and after refinement: cross-attention ($56.3$), self-attention ($43.9$), and other layers (\eg convolution) ($12.4$). The cross-attention layer had the highest importance.
Similar experiments by Custom Diffusion~\cite{kumari2023multi} and E4T~\cite{gal2023e4t} focused on the denoising UNet within LDMs.

\begin{figure}[t]
    \centering
    \includegraphics[width=0.95\linewidth]{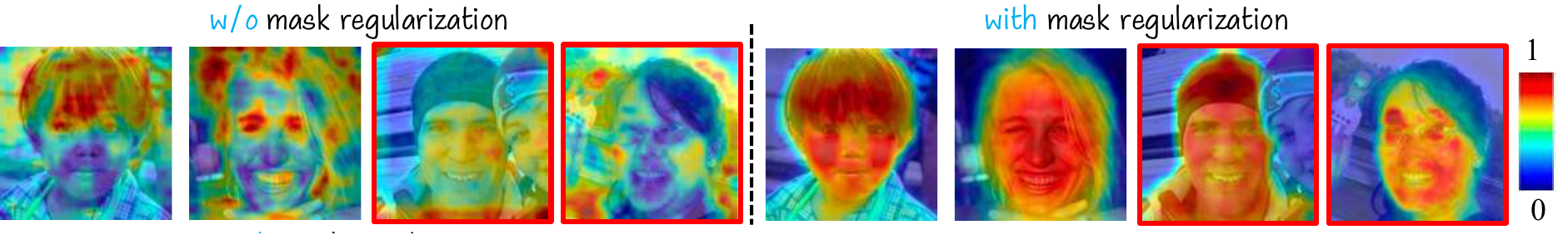}
    \caption{
    Showcases of the cross-attention map for the subject word `S*'. Applying mask regularization $\mathcal{L}_{\boldsymbol{M}}$ prevents attention leakage to irrelevant areas, even with other subjects present (highlighted in red boxes).
    }
    \label{fig:effect_mask}
\end{figure}

\paragraph{Training with Mask Regularization.} The diffusion loss $\mathcal{L}_{\boldsymbol{\epsilon}}$ (\cref{eq:ldmloss}) is commonly used for training the prompt regressor. However, as shown in \cref{fig:effect_mask} (Left), the cross-attention maps of the subject's word embedding often leak into the irrelevant background. 
To address this, we use the subject's mask to constrain the cross-attention maps of subject words, ensuring focus on the subject region. Let $\boldsymbol{A}_{\boldsymbol{e}_i} \in \mathbb{R}^{h\times w}$ be the attention map for $\boldsymbol{e}_i$, and $\boldsymbol{M} \in \mathbb{R}^{H\times W}$ be the segmentation mask for the subject. We use a mask regularization loss to align $\boldsymbol{A}_{\boldsymbol{e}_i}$ with $\boldsymbol{M}$:
\begin{align}
    \mathcal{L}_{\boldsymbol{M}} = \frac{1}{n} \sum_{i=1}^{n}  \text{mean}\left(\boldsymbol{A}_{\boldsymbol{e}_i} \cdot (\boldsymbol{1} - \boldsymbol{M})\right) - \text{mean}(\boldsymbol{A}_{\boldsymbol{e}_i} \cdot \boldsymbol{M}), 
    \label{eq:maskloss}
\end{align}
where $n$ is the number of word embeddings. This loss minimizes attention outside the mask and maximizes it inside, ensuring focus on the subject. The interpolation of $\boldsymbol{M}$ to match $\boldsymbol{A}_{\boldsymbol{e}_i}$ is omitted for simplicity.
The final training objective of the regressor is:
\begin{align}
    \mathcal{L} = \mathcal{L}_{\boldsymbol{\epsilon}} + \alpha_{\boldsymbol{M}} \mathcal{L}_{\boldsymbol{M}},
    \label{eq:totalloss}
\end{align}
where $\alpha_{\boldsymbol{M}}$ is the weight of $\mathcal{L}_{\boldsymbol{M}}$. During refinement, $\mathcal{L}$ is used to prevent diverging update directions, focusing on fine-tuning the cross-attention parameters.

%% file: camera_ready/exp.tex
\section{Experiments}

\subsection{Experiemental Setup}

\paragraph{Datasets.}
We mainly conduct experiments on facial images. Following prior research practices, we train~\name~on the aligned FFHQ dataset~\cite{ffhq}, comprising $70,000$ in-the-wild facial images. 
Constrained by computational resources, we conducted testing on the first $1,000$ images from the CelebA-HQ test split.
In addition, we also evaluate \name~on other categories to demonstrate its versatility. 
With the same setting of ELITE~\cite{wei2023elite}, we train \name~on the test split of OpenImagesv6~\cite{krasin2017openimages}, and test on the DreamBooth Dataset. 
All images are cropped and resized to $512\times 512$ pixels.

\paragraph{Training Details.} 
We use Stable Diffusion v$1.5$~\cite{SD} as our Text-to-Image model, initializing the prompt regressor with the UNet from Stable Diffusion. The DINOv$2$-Large model~\cite{oquab2023dinov2} extracts the DINO feature, and the vision model of clip-vit-large-patch14~\cite{clip} extracts the CLIP image feature.
The segmentation mask $\boldsymbol{M}$ (\cref{eq:maskloss}) is obtained using InSPyReNet~\cite{kim2022revisiting}. We use the same hyper-parameters for both facial and non-facial experiments.
For the word embedding probe stage, we use the AdamW optimizer~\cite{loshchilov2018decoupled} with a learning rate of $2\mathrm{e}{-5}$ and a batch size of $8$. Training on a single Nvidia A$100$ GPU takes $40$ hours.
In the refinement stage, we fine-tune the cross-attention parameters of the regressor for $5$ steps using AdamW, with a learning rate of $2\mathrm{e}{-5}$ and a weight decay of $1\mathrm{e}{-2}$. Parameters $\alpha_{\boldsymbol{M}}$ (\cref{eq:totalloss}) and $\lambda$ (\cref{eq:partial_refine}) are set to $1\mathrm{e}{-3}$ and $1\mathrm{e}{-2}$, respectively.

\begin{figure}[t]
    \centering
    \includegraphics[width=0.9\linewidth]{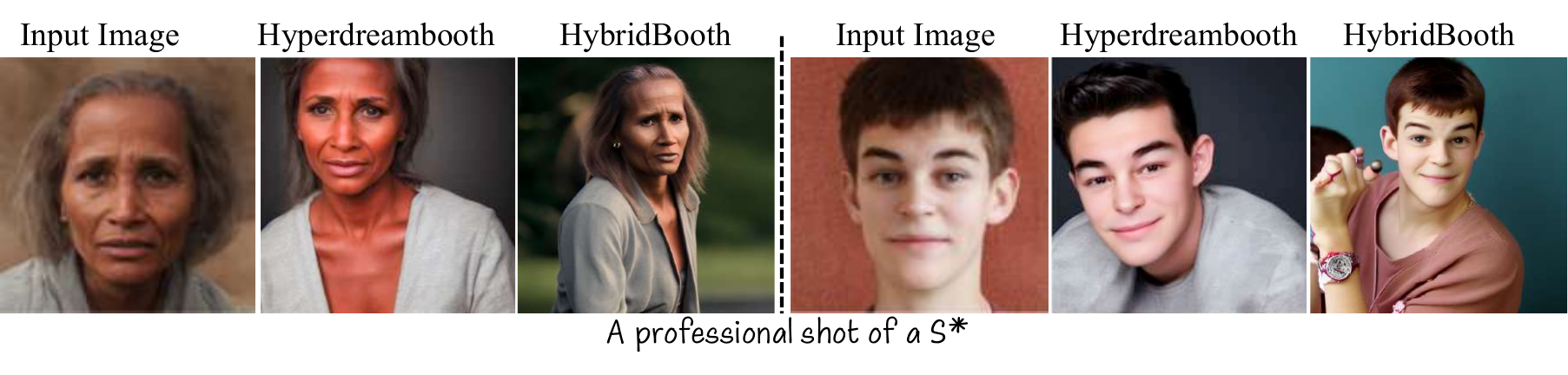}
    \caption{Qualitative Comparison with Hyperdreambooth~\cite{ruiz2023hyperdreambooth}.}
    \label{fig:with_hyperdb}
\end{figure}

\paragraph{Baselines.} We compare the proposed \name~with two categories of methods:
($1$) Optimization-based methods: DreamBooth~\cite{dreambooth}, Textual Inversion~\cite{textualinversion}, and Custom Diffusion~\cite{kumari2022customdiffusion}.
($2$) Direct-regression-based methods: ELITE~\cite{wei2023elite} and FastComposer~\cite{xiao2023fastcomposer}.
Optimization-based methods typically require $3$$\sim$$5$ images as input, which are not available in CelebA-HQ (our test set). To ensure a fair comparison, we augment each image to generate $5$ images for these methods using affine transformations and background changes. More details on the augmentation process are provided in the Supplementary.
To clarify the difference between \name~and HyperDreamBooth~\cite{ruiz2023hyperdreambooth}, HyperDreamBooth updates weights in a low-rank space for rapid model tuning but overlooks the importance of training a powerful encoder initially. Since the official code for HyperDreamBooth was unavailable during our submission, we could only make qualitative comparisons with examples from their paper (see \cref{fig:with_hyperdb}).

\paragraph{Evaluation Metrics.} \label{sec:exp_detail}
We evaluate the model's performance in two aspects:
\begin{itemize}
    \item \textbf{Prompt Fidelity}: This measures how well the generated images match the textual prompt, quantified by the average cosine similarity of CLIP embeddings between the prompt and the generated images ({CLIP-T}). \textit{Higher CLIP-T indicates better textual editability.}
    \item \textbf{Subject Fidelity}: This assesses the preservation of subject details during generation. We use {CLIP-I} and {DINO-I} metrics, which are the cosine similarities between the embeddings of the generated images and the original concept images. \textit{Higher values indicate better subject fidelity.}
\end{itemize}

\begin{figure}[t]
    \centering
    \includegraphics[width=0.9\linewidth]{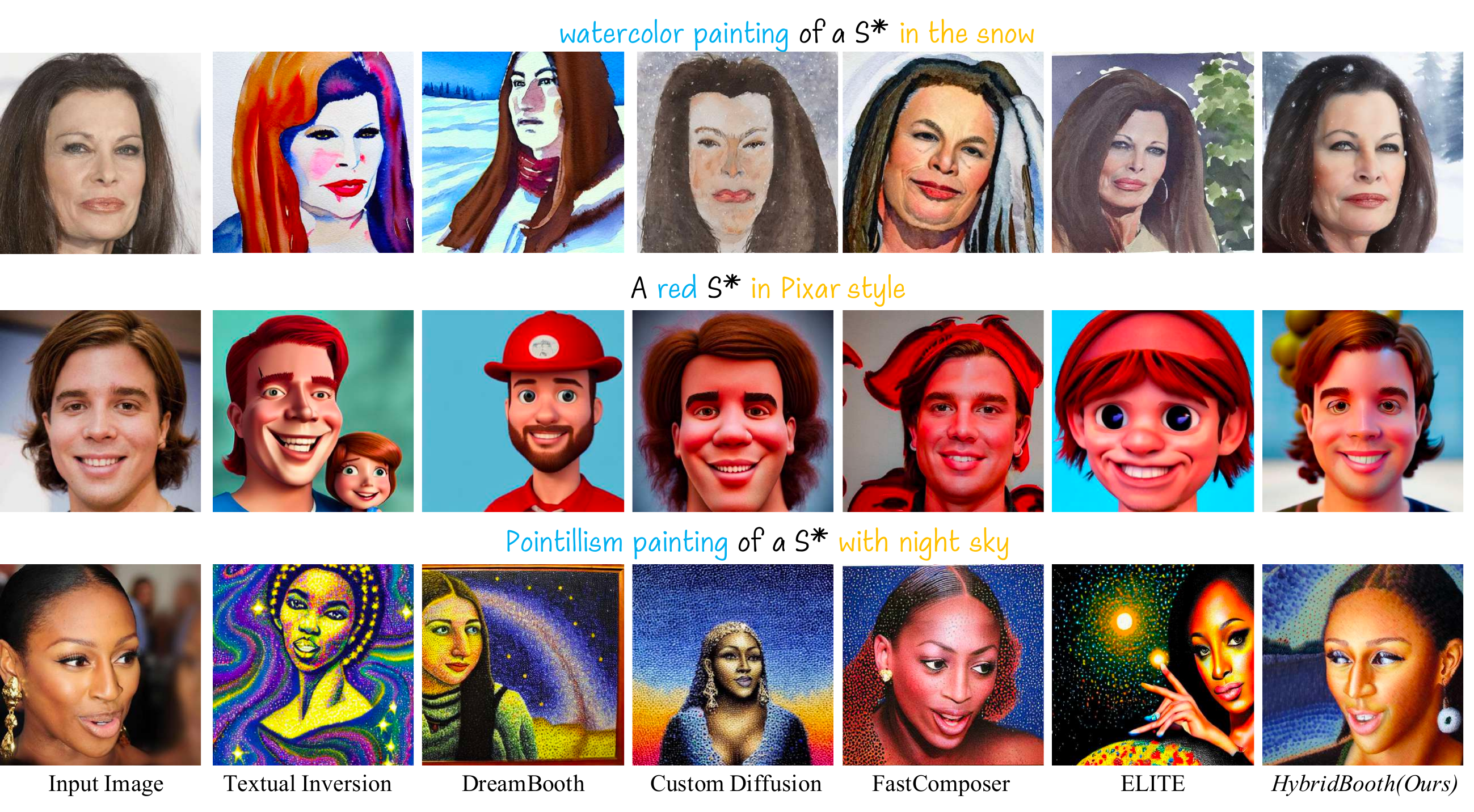}
    \caption{Qualitative Comparison Results on CelebA-HQ (facial dataset). Columns $2$-$4$: Optimization-based methods. 
    Columns $5$-$6$: Direct-regression-based methods.}
    \label{fig:comparison}
\end{figure}

\begin{figure}[t]
    \centering
    \includegraphics[width=0.9\linewidth]{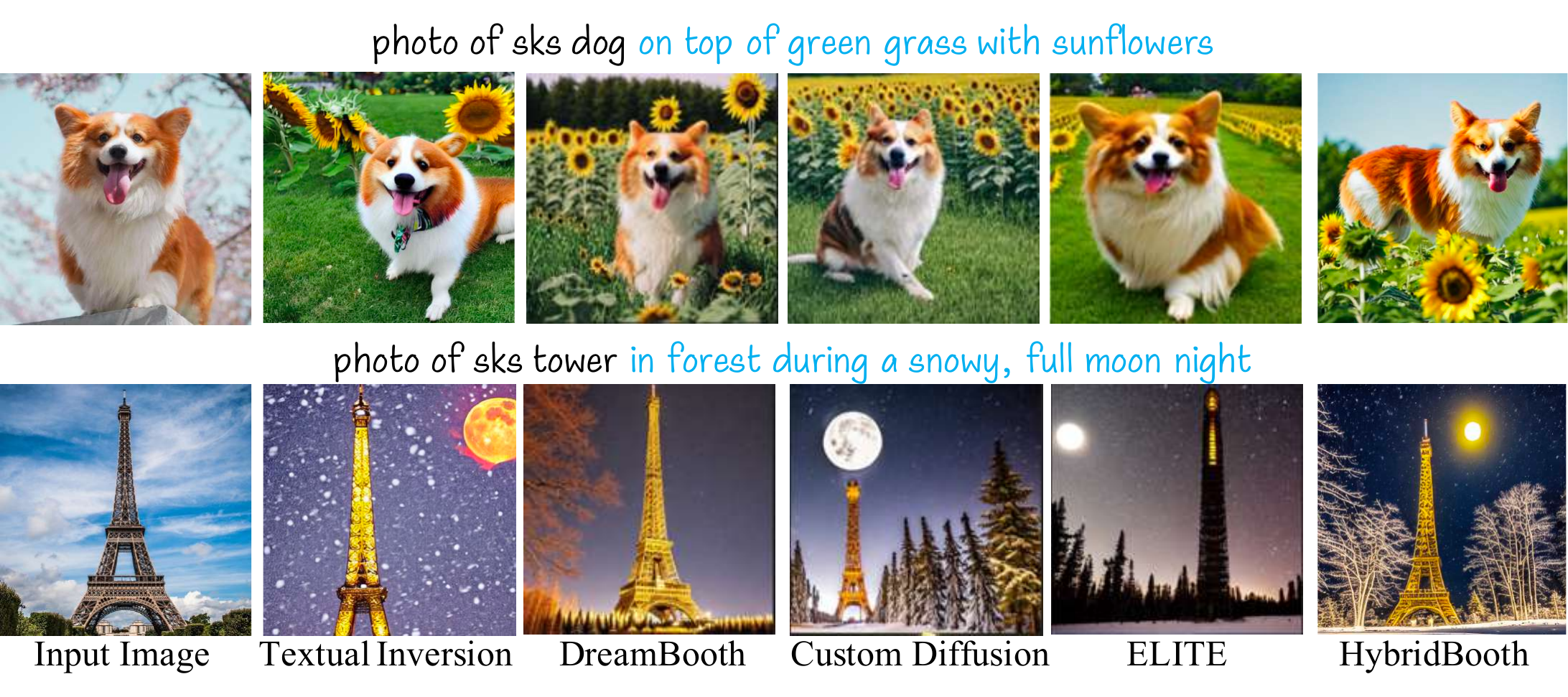}
    \caption{Qualitative Comparison Results on non-facial images. }
    \label{fig:general_obj}
\end{figure}

\subsection{Qualitative Comparison}
In \cref{fig:comparison}, we present qualitative comparison results. Similar to previous works, `S*' denotes the word embedding learned from the input subject image. For a fair comparison, optimization-based methods use 5 images for optimization. Our setting is more complex, specifying both artistic style (e.g., 'watercolor painting') and environmental characteristics (e.g., 'in the snow'). From the results, two key observations emerge:
(1) Optimization-based methods (Columns 2-4), using multiple image inputs and finely tuned hyper-parameters (e.g., total iteration steps), generally outperform direct-regression-based methods (Columns 5-6) in prompt alignment and subject similarity.
(2) Our proposed \name~outperforms all compared methods in text alignment and subject similarity. For example, in the third row, other methods either fail in text similarity (e.g., FastComposer) or in preserving facial similarity, whereas \name~successfully aligns the subject with the editing prompt, demonstrating its superiority.

Qualitative comparisons with HyperDreamBooth~\cite{ruiz2023hyperdreambooth} are shown in \cref{fig:with_hyperdb}. HyperDreamBooth trains a hyper-network to predict low-rank weights of the image diffusion model and fine-tunes these weights on one subject image. However, low-rank weight updates limit personalized representation, resulting in poorer subject feature preservation, as seen in \cref{fig:with_hyperdb}. This supports our core idea that a more powerful encoder reduces the cost of iterative refinement.

As shown in \cref{fig:general_obj}, we evaluate \name~on non-facial images, including animals (Line 1) and complex buildings (Line 2). FastComposer~\cite{xiao2023fastcomposer} is excluded as it is designed for facial images. In Line 1, most methods preserve the texture of the subject well. However, in Line 2, only \name~maintains the structural information while adhering to the complex prompt. This is due to the hybrid framework, which enables quick feature learning while retaining the prompt understanding capability of the Stable Diffusion model.

\begin{table}[t]
    \centering
    \caption{Quantitative Evaluation on the CelebA-HQ. \textbf{First Block}: Optimization-based methods. \textbf{Second Block}: Direct-regression-based methods. \textbf{Third Block}: ~\name~ and its ablations.}
    \resizebox{0.95\linewidth}{!}{
    \begin{tabular}{l|ccc|ccc|c}
    \toprule
        &\multicolumn{3}{c|}{CelebA-HQ} &\multicolumn{3}{c|}{DreamBooth-dataset}  \\
        Method              &CLIP-T $^\uparrow$  &CLIP-I $^\uparrow$ &DINO-I $^\uparrow$ &CLIP-T $^\uparrow$  &CLIP-I $^\uparrow$ &DINO-I $^\uparrow$ &Iter. Step$^\downarrow$  \\
    \midrule 
        Textual Inversion~\cite{textualinversion}               &0.164&0.612&0.236 &0.183&0.663&0.462 &5000 \\
        DreamBooth~\cite{dreambooth}                            &\textbf{0.251}&0.564&0.376 &0.251&{0.785}&0.674 &~1000 \\
        Custom Diffusion~\cite{kumari2022customdiffusion}       &0.237&0.675&0.398 &0.245&0.801&0.695 &200 \\
    \midrule
        ELITE~\cite{wei2023elite}                               &0.169&0.592&0.311 &\underline{0.255}&0.762&0.652 &1\\
        FastComposer~\cite{xiao2023fastcomposer}                &0.201&0.782&\underline{0.581} &-&-&- &1\\
    \midrule
        \name~                           &\underline{0.246}&\textbf{0.865} &\textbf{0.644} &\textbf{0.261}&\textbf{0.865}&\textbf{0.755} &5 \\
        - \textit{w/o} Refinement                               &0.177&\underline{0.842}&{0.568} &0.192&0.835&0.645 &-\\
        - \textit{w/o} Probe                                    &0.153&0.408&0.068 &0.142&0.412&0.050 &5\\
        - {\textit{w/o} DINO Feature }                            &0.161&0.837&0.453 &0.184&\underline{0.841}&0.596 &5  \\
        - {\textit{w/o} CLIP Feature }                            &0.182&0.734&0.510  &0.206&0.752&0.630 & 5  \\
        - {\textit{w/o} Mask Regularization}                      &0.203&0.831&0.625  &0.233&0.833&\underline{0.705} & 5 \\
    \bottomrule
    \end{tabular}
    }
    \label{tab:quantres}
\end{table}

\subsection{Quantitative Comparison}
The quantitative results in \cref{tab:quantres} include the necessary iteration steps for assessing time efficiency. Optimization-based methods generally outperform direct-regression-based methods but require more iteration steps. Our~\name~significantly exceeds other methods in both prompt fidelity and subject fidelity. Notably, our DINO-T metric is over 10\% higher than FastComposer, a key metric in the DreamBooth framework~\cite{dreambooth}.

\begin{figure}[t]
    \centering
    \includegraphics[width=0.92\linewidth]{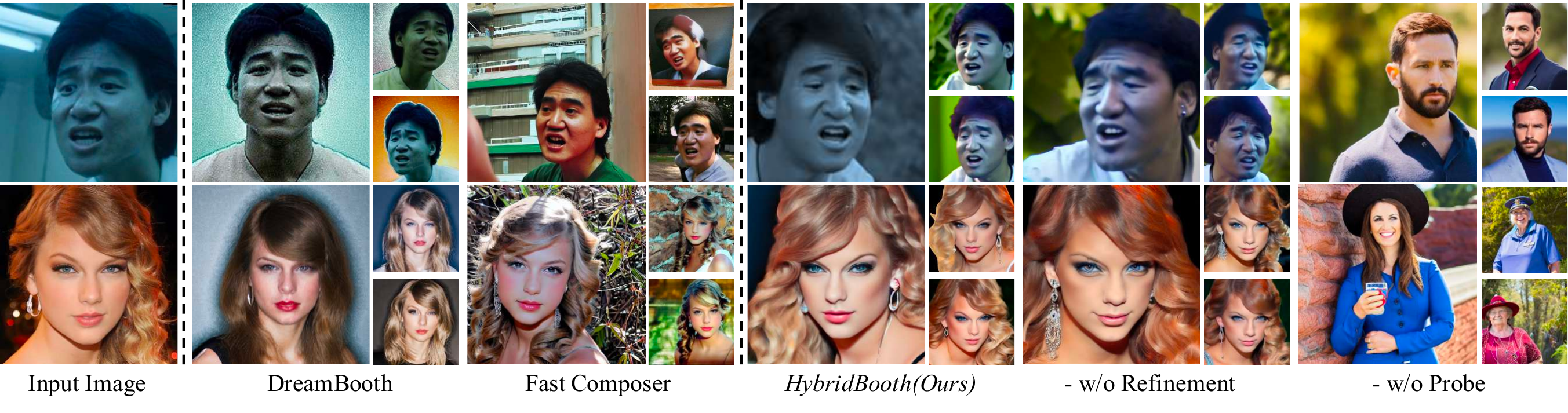}
    \caption{We conducted ablations on~\name~to evaluate its influence on learning subject-specific word embeddings. 
    DreamBooth and FastComposer are utilized here for comparison. 
    `\textit{A photo of S*}' is used for generating images.}
    \label{fig:framework_ablation}
\end{figure}

\begin{figure}[t]
    \centering
    \includegraphics[width=0.85\linewidth]{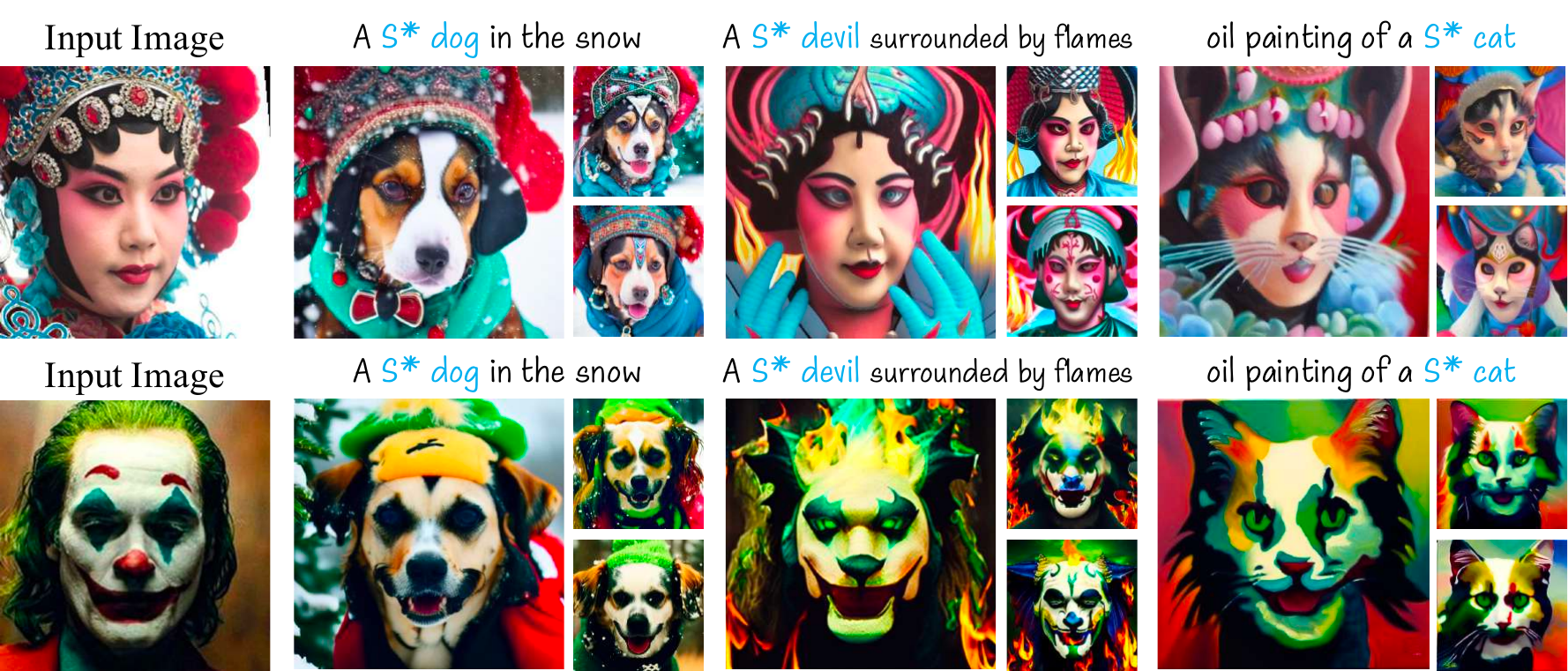}
    \caption{The learned words from human beings can be applied to cross-species subjects, which verifies that~\name~manages to learn intrinsic subject features. 
    }
    \label{fig:crossbeings}
\end{figure}

 \begin{figure}[t]
    \centering
    \includegraphics[width=0.75\linewidth]{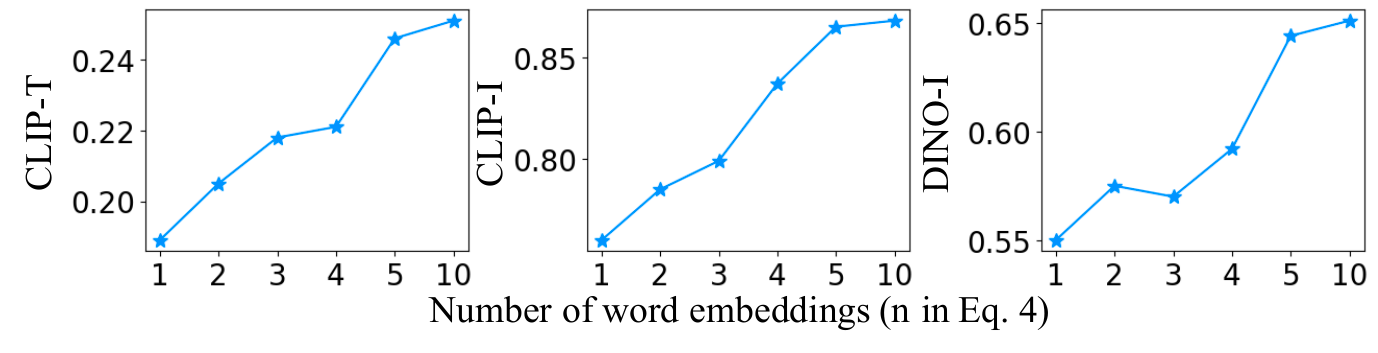}
    \caption{Impact of increasing number of word embeddings (on CelebeA-HQ).}
    \label{fig:wordembed_num}
\end{figure}

\subsection{Ablation Analysis}\label{sec:analysis}
\paragraph{Ablations on the Hybrid Framework.}
We evaluate our hybrid framework, which includes a word embedding stage for initial estimation and a residual refinement stage for precise alignment. Ablations, termed '\textit{w/o} Probe' and '\textit{w/o} Refinement', are shown in \cref{fig:framework_ablation} and \cref{tab:quantres}. The prompt regressor is initialized using the pre-trained UNet from Stable Diffusion in both settings. We assess subject fidelity using the prompt `A photo of S*,' comparing it with DreamBooth and FastComposer.
As shown in \cref{fig:framework_ablation}, subject embedding from residual refinement alone leads to lower subject similarity. Without refinement, word embedding only crudely captures key features. Our~\name~framework, integrating both probe and refinement stages, outperforms all ablations and compared methods, effectively restoring subject characteristics.

\paragraph{Ablations on Multi-grained Image Feature Merging.}
Previous works~\cite{xiao2023fastcomposer,wei2023elite} typically use CLIP features as input for the regressor. As discussed in \cref{sec:probe}, we use CLIP features for global semantic information and DINO features for detailed pixel-level information. We ablate the feature extraction strategy to verify this approach.
As shown in \cref{tab:quantres}, relying solely on DINO or CLIP features results in suboptimal performance. However, integrating both features significantly enhances performance. Notable improvements in prompt fidelity and subject fidelity are observed, even with just a single layer normalization and linear layer.

\paragraph{Analysis on the Mask Regularization Loss $\mathcal{L}_{\boldsymbol{M}}$.}
We propose a mask regularization loss $\mathcal{L}_{\boldsymbol{M}}$ (\cref{eq:maskloss}) to ensure that the estimated word embeddings align with the subject area. As shown in the last line of \cref{tab:quantres}, both subject fidelity and prompt fidelity decrease significantly without the mask regularization loss.

\paragraph{Ablations on the Number of Word Embedding.}Recalling that the prompt regressor estimates multiple word embeddings for enhanced description of key subject features, we adjust the number of embeddings. Quantitative results in \cref{fig:wordembed_num} indicate a consistent improvement in metrics related to Prompt Fidelity and Subject Fidelity with an increasing number of words. To balance efficiency and computation cost, we set the number of word embeddings we estimate to $5$.

\paragraph{Can the Learned Subject Concept be Cross-species Adapted?} 
The well-aligned training data in the \textit{Probe Stage} results in the encoder developing a consistent semantic tendency in attention maps, as shown in \cref{fig:effect_mask}. The subsequent \textit{Refinement Stage} further enhances the model's generalization ability. \cref{fig:crossbeings} illustrates that species with similar semantic structures, such as humans, dogs, and cats, consistently yield robust results, even with complex prompts.

%% file: camera_ready/concl.tex
\begin{figure}[t]
    \centering
    \includegraphics[width=0.75\linewidth]{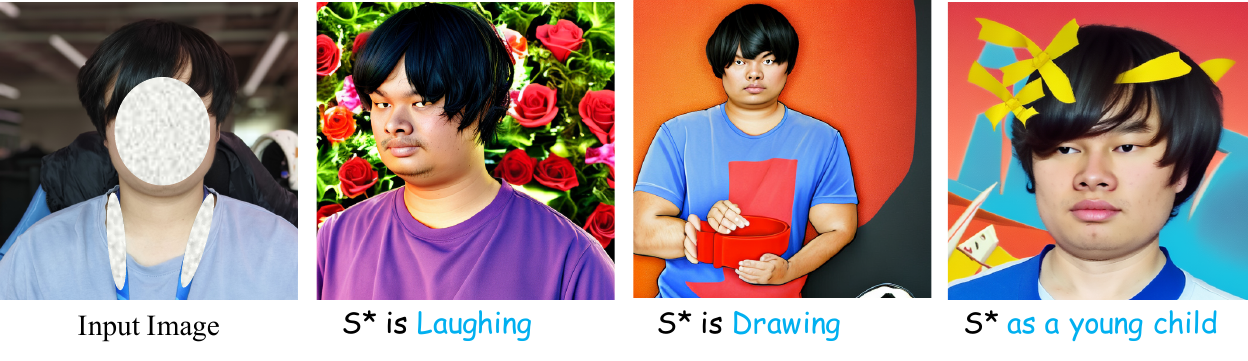}
    \caption{Showcases of limitations in~\name. 
    }
    \label{fig:limitation}
\end{figure}

\section{Conclusion}
In this paper, we address the challenge of achieving fast prompt inversion while preserving the generalization capabilities of text-to-image models. By leveraging both optimization-based and direct-regression-based methods, we propose a two-stage hybrid prompt inversion framework, termed~\name. This framework ensures robust and efficient subject-driven generation while maintaining the model's general creative ability. Experimental results demonstrate the superior performance of~\name~compared to existing state-of-the-art methods.

\noindent \textbf{Limitation.} As shown in \cref{fig:limitation}, \name~cannot perform precise semantic editing, such as adjusting expressions and age. In addition, \name~inherits the weakness from StableDiffusion that fails to generate fine-grained details of the subject, such as fingers. To address these issues, we believe possible solutions are incorporating language-and-vision models~\cite{liu2024improved} for improved understanding and planning, and utilizing more 3D structural information~\cite{bogo2016keep,3dmm}.

\section*{Acknowledgement}
This work was supported in part by the National Natural Science Foundation of China under Grant No. 62073244 and 61825303, 62088101, and the Shanghai Municipal Science and Technology Major Project (2021SHZDZX0100) and the Fundamental Research Funds for the Central Universities. We thank Yixuan Li for his assistance with the illustrations. 